\documentclass[11pt,a4paper]{article}
\usepackage[hyperref]{acl2020}
\usepackage{times}
\usepackage{latexsym}

\usepackage{balance}

\usepackage{bm}
\usepackage{amssymb}
\usepackage{amsmath}
\usepackage{url}
\usepackage{makecell}
\usepackage{multirow}
\usepackage{subfigure}
\usepackage{graphicx} 
\usepackage{color}
\usepackage{colortbl}
\usepackage{textcomp}
\usepackage{CJKutf8}
\usepackage{booktabs}
\usepackage{microtype}
\usepackage{arydshln}
\usepackage{xspace}

\aclfinalcopy 
\def\aclpaperid{393} 

\newcommand{\eg}{\emph{e.g.,}\xspace}

\title{Self-Attention with Cross-Lingual Position Representation}

\author{Liang Ding$^\dagger$~~~~~Longyue Wang$^\ddagger$~~~~~Dacheng Tao$^{\dagger}$
\\
  $^\dagger$UBTECH Sydney AI Centre, School of Computer Science,\\Faculty of Engineering, The University of Sydney\\
  {\tt \{ldin3097,dacheng.tao\}@sydney.edu.au} \\
  $^\ddagger$Tencent AI Lab\\
  {\tt vinnylywang@tencent.com}}

\date{}

\begin{document}
\maketitle
\begin{abstract}
Position encoding (PE), an essential part of self-attention networks (SANs), is used to preserve the word order information for natural language processing tasks, generating fixed position indices for input sequences. However, in cross-lingual scenarios, \eg machine translation, the PEs of source and target sentences are modeled independently. Due to word order divergences in different languages, modeling the cross-lingual positional relationships might help SANs tackle this problem.
In this paper, we augment SANs with \emph{cross-lingual position representations} to model the bilingually aware latent structure for the input sentence. Specifically, we utilize bracketing transduction grammar (BTG)-based reordering information to encourage SANs to learn bilingual diagonal alignments. Experimental results on WMT'14 English$\Rightarrow$German, WAT'17 Japanese$\Rightarrow$English, and WMT'17 Chinese$\Leftrightarrow$English translation tasks demonstrate that our approach significantly and consistently improves translation quality over strong baselines. Extensive analyses confirm that the performance gains come from the cross-lingual information.

\end{abstract}

\begin{figure}[tb]
    \centering
    \includegraphics[width=0.43\textwidth]{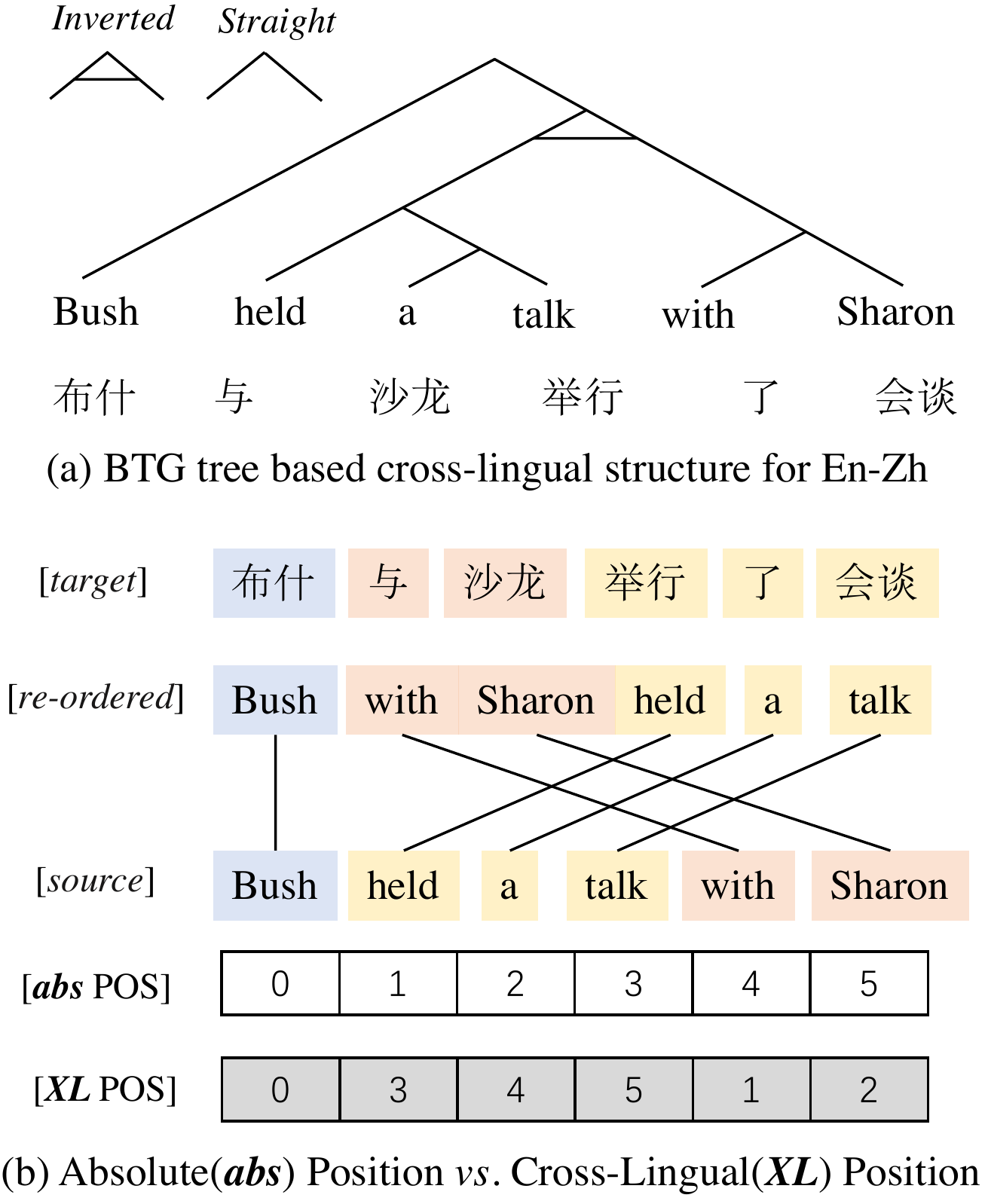}
    \caption{Illustration of cross-lingual position for English$\Rightarrow$Chinese translation task. (a) BTG tree shows the cross-lingual preordering. The top-left corner is the transduction grammar. (b) the difference between absolute position encoding (APE) 
    and our proposed cross-lingual position encoding (XL PE)
    .}
    \label{fig:BTG_example}
\end{figure}

\section{Introduction}
Although self-attention networks (SANs)~\cite{lin2017structured} have achieved the state-of-the-art performance on several natural language processing (NLP) tasks~\cite{transformer,devlin2019bert,radford2018improving}, they possess the innate disadvantage of sequential modeling due to the lack of positional information. Therefore, absolute position encoding (APE)~\cite{transformer} and relative position encoding (RPE)~\cite{shaw2018self} were introduced to better capture the sequential dependencies. 
However, either absolute or relative PE is language-independent and its embedding remains fixed. This inhibits the capacity of SANs when modelling multiple languages, which have diverse word orders and structures~\cite{GellMann17290}.
Recent work have shown that modeling cross-lingual information (\eg alignment or reordering) at 
encoder or attention level
improves translation performance for different language pairs \cite{cohn2016incorporating,du2017pre,zhao2018exploiting,kawara2018recursive}.

Inspired by their work, we propose to augment SANs with \emph{cross-lingual representations}, by encoding reordering indices at embedding level. Taking English$\Rightarrow$Chinese translation task for example, we first reorder the English sentence by deriving a latent bracketing transduction grammar (BTG) tree~\cite{wu1997stochastic} (Fig.~\ref{fig:BTG_example}a). Similar to absolute position, the reordering information can be represented as cross-lingual position (Fig.~\ref{fig:BTG_example}b). In addition, we propose two strategies to incorporate cross-lingual position encoding into SANs. 
We conducted experiments on three commonly-cited datasets of machine translation. Results show that exploiting cross-lingual PE consistently improves translation quality
. Further analysis reveals that our method improves the alignment quality ({\S}Sec.~\ref{subsec:aq}) and 
context-free Transformer
~\cite{tang2019understanding} ({\S}Sec.~\ref{subsec:encoder-free}). Furthermore, contrastive evaluation demonstrates that NMT models benefits from the cross-lingual information rather than denoising ability ({\S}Sec.~\ref{subsec:contrastivee_val}).

\section{Background}
\paragraph{Position Encoding}
To tackle the position unaware problem, absolute position information 
is injected into the SANs:
\begin{equation}
    \mathbf{PE}_{abs}=f(pos_{abs}/10000^{2i/d_{model}}) \label{eq:pos}
\end{equation}
where $pos_{abs}$ denotes the numerical position indices, $i$ is the dimension of the position indices and $d_{model}$ means hidden size. 
$f(\cdot)$ alternately employs $sin(\cdot)$ and $cos(\cdot)$ for even and odd dimensions. Accordingly, the position matrix $\mathbf{PE}$ can be obtained
given the input $\mathbf{X} = \{\mathbf{x}_1, \dots, \mathbf{x}_T\} \in \mathbb{R}^{T \times d_{model}}$. Then, the position aware output $\mathbf{Z}$ is calculated by:
\begin{equation}
    \mathbf{Z}=\mathbf{X}+\mathbf{PE}_{abs} \qquad  \in \mathbb{R}^{T\times d_{model}} \label{eq:add}
\end{equation}
\paragraph{Self-Attention}
The SANs 
compute the attention of each pair of elements in parallel. 
It first converts the input into three matrices $\mathbf{Q}, \mathbf{K}, \mathbf{V}$, representing queries, keys, and values, respectively:
\begin{equation}
    \{\mathbf{Q},\mathbf{K},\mathbf{V}\}=\\
    \{\mathbf{Z}\mathbf{W}_Q,\mathbf{Z}\mathbf{W}_K,\mathbf{Z}\mathbf{W}_V\}
    \label{eq:input}
\end{equation}
where $\mathbf{W}_Q, \mathbf{W}_K, \mathbf{W}_V \in \mathbb{R}^{d_{model} \times d_{model}}$ are parameter matrices.
The output is then computed as a weighted sum of values by $\textsc{Att}(\mathbf{Q,K,V})$.
SANs can be implemented with multi-head attention mechanism, which requires extra splitting and concatenation operations. Specifically, $\mathbf{W}_Q, \mathbf{W}_K, \mathbf{W}_V$ and $\mathbf{Q}, \mathbf{K}, \mathbf{V}$ in Eq.~(\ref{eq:input}) is split into H sub-matrices, yielding H heads. For the $h$-th head, the output is computed by:
\begin{equation}
    \mathbf{O}_h = \textsc{Att}(\mathbf{Q}_h,  \mathbf{K}_h, \mathbf{V}_h) \in \mathbb{R}^{T \times d_{v}} \label{eq:output}
\end{equation}
Where subspace parameters are $\mathbf{W}_Q^{h}, \mathbf{W}_K^{h}\in\mathbb{R}^{d_{model}\times{d_k}}$ and $\mathbf{W}_V^{h}\in\mathbb{R}^{d_{model}\times{d_v}}$, where $d_k, d_v$ refer to the dimensions of keys and values in the subspace, and normally $d_k=d_v=d_{model}/\textsc{H}$. Finally, these subspaces are combined with concatenation operation:
\begin{equation}
    \mathbf{O}=\textsc{concat}(\mathbf{O}_1,\dots,\mathbf{O}_H)\mathbf{W}_O \label{eq:concat}
\end{equation}
where $\mathbf{W}_O \in \mathbb{R}^{Hd_v \times d_{model}}$ and $\mathbf{O} \in \mathbb{R}^{T \times d_{model}}$ are the parameter matrix and output, respectively.

\section{Approach}
\subsection{Cross-Lingual Position Representation}
First, we built a BTG-based reordering model~\cite{neubig2012inducing} to generate a reordered source sentence according to the word order of its corresponding target sentence. 
Second, we obtained the reordered word indices $pos_{\textsc{xl}}$ that correspond with the input sentence $\mathbf{X}$. To output the cross-lingual position matrix $\mathbf{PE}_{\textsc{xl}}$, we inherit the sinusoidal function in Eq. (\ref{eq:pos}). Formally, the process is:
\begin{equation}
    \mathbf{PE}_{\textsc{xl}}=f(\textsc{BTG}(\mathbf{X}))
    \label{eq:pos_xl}
\end{equation}

\begin{figure}[tb]
    \centering
    \includegraphics[width=0.48\textwidth]{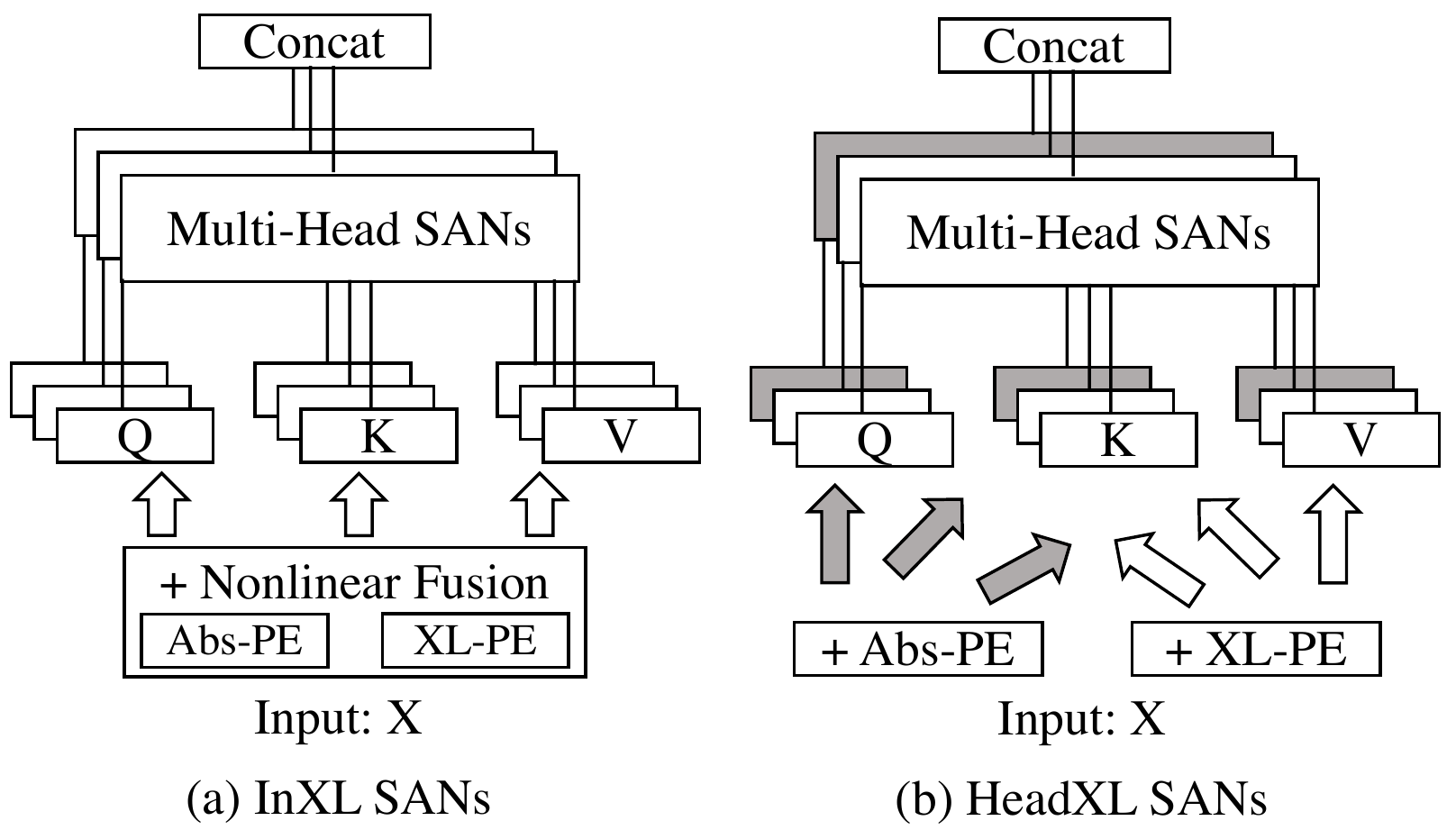}
    \caption{The proposed integration strategies.}
    \label{fig:structure}
\end{figure}

\subsection{Integration Strategy}
As shown in Fig.~\ref{fig:structure}, we propose two strategies to integrate the cross-lingual position encoding (XL PE) into SANs: inputting-level XL (\textbf{InXL}) SANs and head-level (\textbf{HeadXL}) SANs. 
\paragraph{Inputting-level XL SANs}
As illustrated in Fig.~\ref{fig:structure}a, we employ a non-linear function $\textsc{Tanh}(\cdot)$ to fuse $\mathbf{PE}_{abs}$ and $\mathbf{PE}_{\textsc{xl}}$:
\begin{equation}
    \mathbf{PE}_{\textsc{In-XL}}=\textsc{Tanh}(\textbf{PE}_{abs}\mathbf{U}+\textbf{PE}_{\textsc{xl}}\mathbf{V}) \label{eq:InXL}
\end{equation}
where $\mathbf{U},\mathbf{V}$ are trainable parameters. {In our preliminary experiments, the non-linear function performs better than element-wise addition. This might because complex non-linear one have better fitting capabilities, thereby avoiding exceptional reordering to some extent.}
Next, we perform Eq.~(\ref{eq:add}) to obtain the output representations:
\begin{equation}
    \mathbf{Z}_{\textsc{In-XL}}=\mathbf{X}+\mathbf{PE}_{\textsc{In-XL}} \label{eq:ZInXL}
\end{equation}
Similarly, we use Eq.~(\ref{eq:input})$\sim$(\ref{eq:concat}) to calculate multiple heads of SANs.

\paragraph{Head-level XL SANs}
Instead of projecting XL PE to all attention heads, we feed partial of them, such that some heads contain XL PE and others contain APE, namely HeadXL.
As shown in Fig.~\ref{fig:structure}b, we fist add APE and XL PE for $\mathbf{X}$, respectively:
\begin{equation}
\begin{split}
    \mathbf{Z}_{abs} =& \mathbf{X} + \mathbf{PE}_{abs}\\
    \mathbf{Z}_{\textsc{xl}} =& \mathbf{X} + \mathbf{PE}_{\textsc{xl}} \label{eq:headxl-input}
\end{split}
\end{equation}
We denote the number of XL PE equipped heads as $\tau \in \{0,\dots,H\}$. To perform the attention calculation, $\mathbf{W}_i$ is divided into $[\mathbf{W}_i^{\textsc{xl}} \in \mathbb{R}^{d_{model} \times {\tau}d_v}; \mathbf{W}_i^{abs} \in \mathbb{R}^{d_{model} \times {(H-\tau)}d_v}]$ for each $i\in{\mathbf{Q},\mathbf{K},\mathbf{V}}$, correspondingly generating two types of $\{\mathbf{Q},\mathbf{K},\mathbf{V}\}$ for XL PE heads and APE heads. According to Eq.~(\ref{eq:output}), the output of each XL PE head is:
\begin{equation}
    \mathbf{O}_h^{\textsc{xl}} = \textsc{Att}(\mathbf{Q}_h^{\textsc{xl}},\mathbf{K}_h^{\textsc{xl}},\mathbf{V}_h^{\textsc{xl}}) \in \mathbb{R}^{T \times d_v} \label{eq:headxl-output}
\end{equation}
As a result, the final output of HeadXL is:
\begin{equation}
\begin{split}
    \textsc{HeadSan}(\textsc{X}) =  &\textsc{Concat}(\mathbf{O}^{\textsc{xl}}_1,\dots,\mathbf{O}^{\textsc{xl}}_{\tau}\\
    &\mathbf{O}^{abs}_{\tau+1},\dots,\mathbf{O}^{abs}_H)\mathbf{W}_O \label{eq:head-concat}
\end{split}
\end{equation}
In particular, $\tau = 0$ refers to the original Transformer~\cite{transformer} and $\tau = H$ means that XL PE will propagate over all attention heads.

\begin{table*}[tb]
    \centering
    \renewcommand\arraystretch{1.1}
    \begin{tabular}{c|l|l|l|r}
         \#&\textbf{System}&\textbf{Architecture}&\textbf{BLEU}&\textbf{\#Param.}\\
         \hline
         \hline
         1&\newcite{transformer}&Transformer \textsc{Big}&28.4&213M\\
         2&\newcite{hao2019modeling}&Transformer \textsc{Big} w/ BiARN&28.98&323.5M\\
         3&\newcite{wang2019self}&Transformer \textsc{Big} w/ Structure PE&28.88&--\\
         4&\newcite{chen2019recurrent}&Transformer \textsc{Big} w/ MPRHead&29.11&289.1M\\
         5&\newcite{chen2019neural}&Transformer \textsc{Big} w/ Reorder Emb&29.11&308.2M\\
         \hline
         6&\multirow{6}{*}{This work}&Transformer \textsc{Big}&28.36&282.55M\\
         7&&~~~+ Relative PE&28.71&+0.06M\\
         8&&~~~+ DiSAN&28.76&+0.04M\\
         \cdashline{3-5}
         9&&~~~+ InXL PE&28.66&+0.01M\\
         10&&~~~+ HeadXL PE&28.72&+0.00M\\
         11&&~~~+ Combination &29.05$^\uparrow$&+0.01M\\
    \end{tabular}
    \caption{Experiments on WMT'14 En-De. ``$\uparrow$''indicates significant difference ($p < 0.01$) from Transformer \textsc{Big}. ``\#Param'' denotes the number of parameters. ``+ Combination'' represents combining \#9 and \#10 methods.}
    \label{tab:result}
\end{table*}

\begin{table}[tb]
    \centering
    \renewcommand\arraystretch{1.1}
    \begin{tabular}{l|c|c|c}
         \textbf{System}&\textbf{JaEn}&\textbf{ZhEn}&\textbf{EnZh}\\
         \hline
         \hline
         \newcite{du2017pre}&25.65&--&--\\
         \newcite{hassan2018achieving}&--&24.20&--\\
         \hline
         Transformer \textsc{Big}&29.22&23.94&33.79\\
         ~~~+ Relative PE&29.62&24.36&34.21\\
         ~~~+ DiSAN& 29.73&24.44&34.31\\
         \hdashline
         ~~~+ InXL PE&29.52&24.44&34.23\\
         ~~~+ HeadXL PE&29.62&24.39&34.20\\
         ~~~+ Combination$^\uparrow$ &29.85&24.71&34.51\\
    \end{tabular}
    \caption{Experiments on Ja-En, Zh-En and En-Zh.}
    \label{tab:further_result}
\end{table}

\section{Experiments}
We conduct experiments on word order-diverse language pairs: 
WMT'14 English$\Rightarrow$German (En-De), WAT'17 Japanese$\Rightarrow$English (Ja-En), and WMT'17 Chinese$\Leftrightarrow$English (Zh-En \& En-Zh). 

For English$\Rightarrow$German, the training set consists of 4.5 million sentence pairs and newstest2013 \& 2014 are used as the dev. and test sets, respectively. BPE with 32K merge operations is used to handle low-frequency words.
For Japanese$\Rightarrow$English, we follow~\citet{morishita2017ntt} to use the first two sections as training data, which consists of 2.0 million sentence pairs. The dev. and test sets contain 1790 and 1812 sentences.
For Chinese$\Leftrightarrow$English, we follow~\citet{hassan2018achieving} to get 20 million sentence pairs.
We develop on devtest2017 and test on newstest2017. We use BLEU~\cite{papineni2002bleu} as the evaluation metric with statistical significance test~\cite{collins2005clause}.


\begin{figure}[tb]
    \centering
    \includegraphics[width=0.45\textwidth]{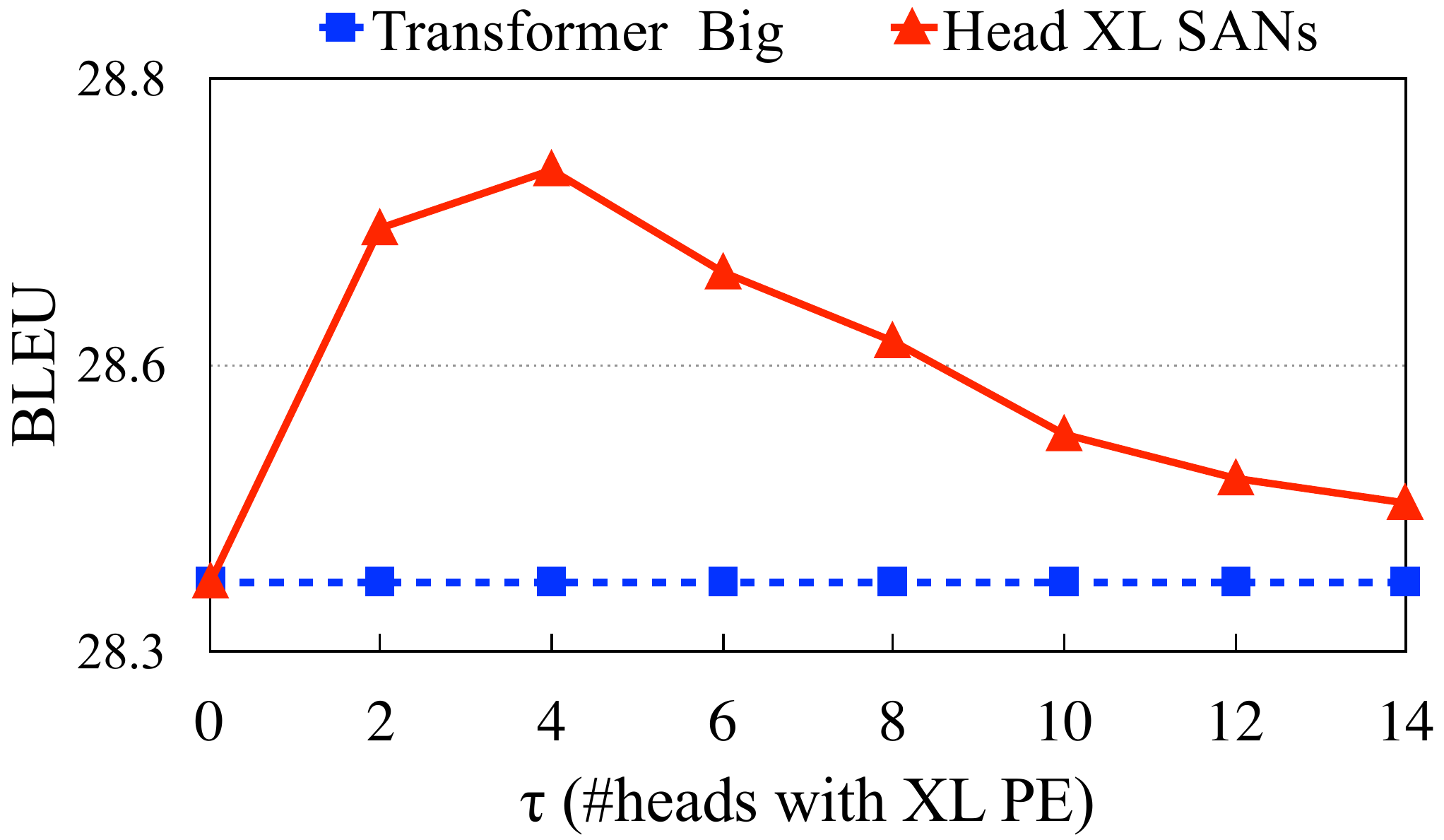}
    \caption{BLEU score on newstest2014 for different $\tau$.}
    \label{fig:ablation_heads}
\end{figure}

We evaluate the proposed XL PE strategies on Transformer. The baseline systems include Relative PE~\cite{shaw2018self} and directional SAN (DiSAN,~\citealt{shen2018disan}). We implement them on top of OpenNMT~\cite{klein2017opennmt}. In addition, we report the results of previous studies~\cite{hao2019modeling,wang2019self,chen2019recurrent,chen2019neural,du2017pre,hassan2018achieving}. 

The reordered source sentences are generated by BTG-based preordering model~\cite{neubig2012inducing}
trained with above sub-word level\footnote{\newcite{garg2019jointly} show that sub-word units are beneficial for statistical model.} parallel corpus. At training phase, we first obtain word alignments from parallel data using GIZA++ or FastAlign, and then the training process is to find the optimal BTG tree for source sentence consistent with the order of the target sentence based on the word alignments and parallel data. At decoding phase, we only provide source sentences as input and the model can output reordering indices, which will be fed into NMT model. Thus, bilingual alignment information is only used to preprocess training data, but not necessary at decoding time.

For fair comparison, we keep the Transformer decoder unchanged and validate different position representation strategies on the encoder. 
We conduct all experiments on the \textsc{Transformer-Big} with four V100 GPUs.

\subsection{Effect of $\tau$ in HeadXL SANs}

Fig.~\ref{fig:ablation_heads} reports the results of different $\tau$ for Head XL \textsc{San}s. With increasing of XL PE-informed heads, the best BLEU is achieved when \#heads = 4, which is therefore left as the default setting for HeadXL. Then, the BLEU score gradually decreases as the number of APE-informed heads decrease ($\tau\uparrow$), indicating that sequential position embedding is still essential for SANs. 

\subsection{Main Results}

Tab.~\ref{tab:result} shows the results on En-De, inputting-level cross-lingual PE (+InXL PE) and head-level cross-lingual PE (+HeadXL PE) outperform Transformer \textsc{Big} by 0.30 and 0.36 BLEU points, and combining these two strategies\footnote{Replace $\mathbf{PE}_\textsc{XL}$ in Eq.~(\ref{eq:headxl-input}) with $\mathbf{PE}_\textsc{In-XL}$ in Eq.~(\ref{eq:ZInXL}).} achieves a 0.69 BLEU point increase. For Ja-En, Zh-En, and En-Zh (Tab.~\ref{tab:further_result}), we observe a similar phenomenon, demonstrating that XL PE on SANs do improve the translation performance for several language pairs. It is worth noting that our approach introduces nearly no additional parameters (+0.01M over 282.55M).

\subsection{Alignment Quality}
\label{subsec:aq}
Our proposed XL PE intuitively encourages SANs to learn bilingual diagonal alignment, so has the potential to induce better attention matrices. We explore this hypothesis on the widely used Gold Alignment dataset\footnote{\url{http://www-i6.informatik.rwth-aachen.de/goldAlignment}, the original dataset is German-English, we reverse it to English-German.}
and follow~\citet{tang2019understanding} to perform the alignment. The only difference being that we average the attention matrices across all heads from the penultimate layer~\cite{garg2019jointly}. The alignment error rate (AER,~\citealt{och2003systematic}), precision (P) and recall (R) are reported as the evaluation metrics. Tab.~\ref{tab:align} summarizes the results. We can see: 1) XL PE allows SANs to learn better attention matrices, thereby improving alignment performance (27.4 / 26.9 vs. 29.7); and 2) combining the two strategies delivers consistent improvements (24.7 vs. 29.7).
\begin{table}[tb]
    \centering
    \renewcommand\arraystretch{1.1}
    \begin{tabular}{l|c|c|c}
         \textbf{Model}&{\textbf{AER}}&\textbf{P}&\textbf{R}  \\
         \hline
         \hline
         Transformer \textsc{Big}&29.7\%&69.9\%&72.7\%\\
         \hline
         ~~+ InXL&27.5\%&72.2\%&74.1\%\\
         ~~+ HeadXL&26.9\%&75.4\%&73.9\%\\
         ~~+ Combination&24.7\%&75.0\%&77.6\%\\
    \end{tabular}
    \caption{The AER scores of alignments on En-De.}
    \label{tab:align}
\end{table}

\subsection{Gain for Context-Free Model}
\label{subsec:encoder-free}
\newcite{tang2019understanding} showed that context-free Transformer (directly propagating the source word embeddings with PE to the decoder) achieved comparable results to the best RNN-based model. We argue that XL PE could further enhance the context-free Transformer. On English$\Rightarrow$German dataset, we compare LSTM-based model, Transformer \textsc{Big}-noenc-nopos, +APE, +RPE and +InXL PE. For fair comparison, we set the LSTM hidden size to 1024. 
In Tab.~\ref{tab:encoder_free}, 
we can see: 1) position information is the most important component for the context-free model, bringing +14.45 average improvement; 2) InXL PE equipped context-free Transformer significantly outperforms the LSTM model while consuming less parameters; and 3) compared to the increment on standard Transformer (+0.30 over 28.36), InXL PE improves more for context-free Transformer (+0.57 over 24.11), where the improvements are +2.3\% vs. +1.1\%.

\begin{table}[tb]
    \centering
    \renewcommand\arraystretch{1.1}
    \begin{tabular}{l|r|r}
         \textbf{System}&\textbf{BLEU}&\textbf{\#Param.}\\
         \hline
         \hline
         LSTM (6 layers)&24.12&178.90M\\
         \hline
         \textsc{Big}-noEnc-noPos&9.97&171.58M\\
         ~~~+ Absolute PE&24.11&+0.00M\\
         ~~~+ Relative PE&24.47&+0.01M\\
         ~~~+ InXL PE&24.68&+0.01M\\
    \end{tabular}
    \caption{Gains over Encoder-Free Transformer.}
    \label{tab:encoder_free}
\end{table}

\subsection{Effects of Noisy Reordering Information}
\label{subsec:contrastivee_val}
\begin{figure}[tb]
    \centering
    \includegraphics[width=0.44\textwidth]{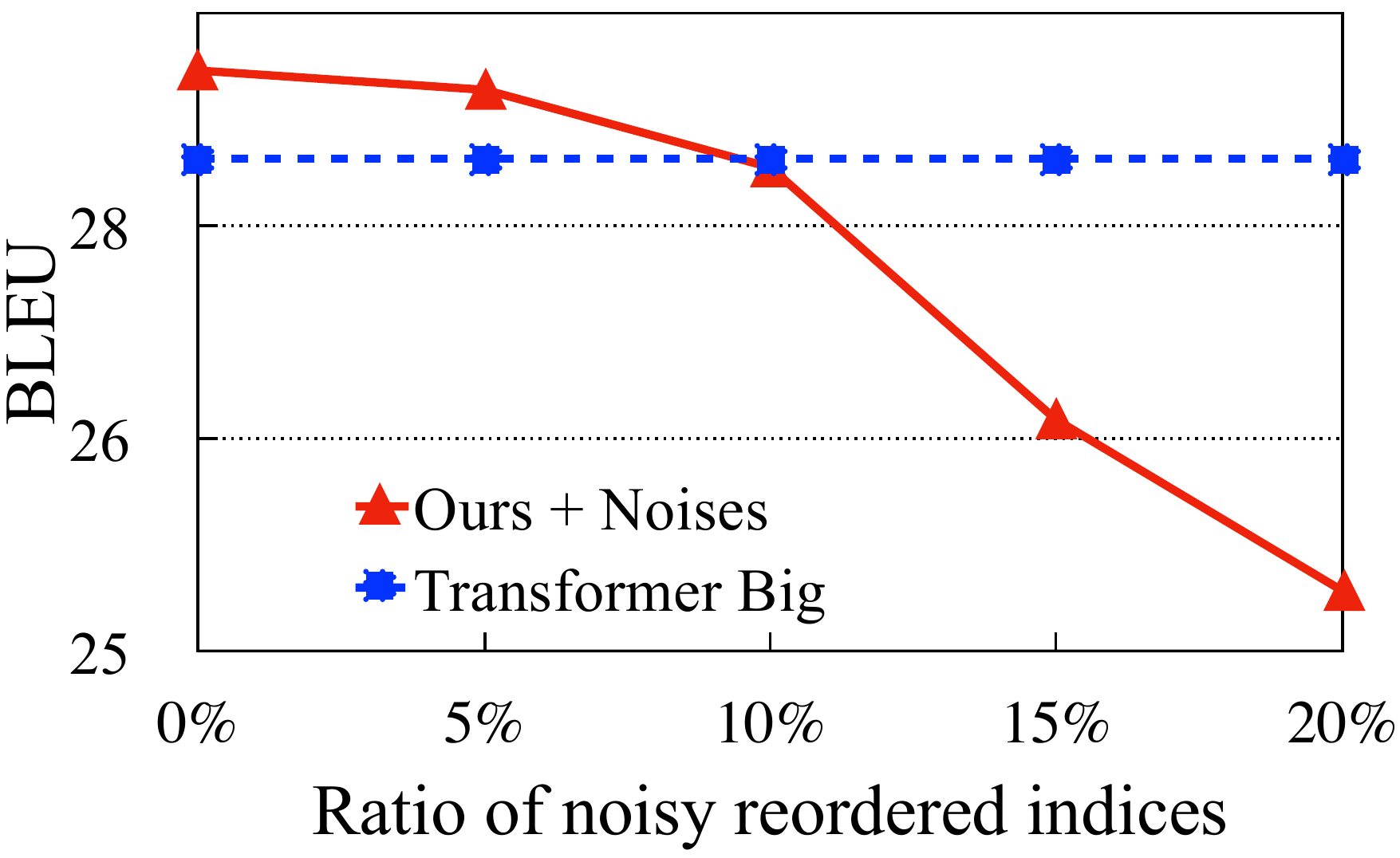}
    \caption{Experiments with noise attacks. Ratio of noisy reordered indices ranges from 0\% to 20\%.}
    \label{fig:noise_attack}
\end{figure}

To demonstrate that our improvements come from cross-lingual position information rather than noisy position signals, 
we attack our model by adding noises\footnote{We randomly swap two reordered positional indexes with different ratios.} into reordered indices of training sentences. As shown in Fig.~\ref{fig:noise_attack}, our method can tolerate partial reordering noises and maintain performance to some extent. However, as noise increases, translation quality deteriorates, indicating that noises in reordering information do not work as regularization. This contrastive evaluation also confirms that the model does not benefit from the noise as much as it benefits from the reordering information.



\section{Related Work}
\paragraph{Augmenting SANs with position representation}
SANs ignore the position of each token due to its position-unaware ``bag-of-words'' assumption. The most straightforward strategy is adding the position representations as part of the token representations~\cite{transformer,shaw2018self}. Besides above sequential PE approaches, \citet{wang2019self} enhanced SANs with structural positions extracted from the syntax dependencies. However, none of them considered modeling the cross-lingual position information between languages.

\paragraph{Modeling cross-lingual divergence}

There has been many works modeling cross-lingual divergence (\emph{e.g.}, reordering) in statistical machine translation~\cite{nagata-etal-2006-clustered,durrani-etal-2011-joint,durrani-etal-2013-markov}. However, it is difficult to migrant them to neural machine translation. 
\citet{kawara2018recursive} pre-reordered the source sentences with a recursive neural network model. \citet{chen2019neural} learned the reordering embedding by considering the relationship between the position embedding of a word and \textsc{Sans}-calculated sentence representation. 
\citet{yang2019assessing} showed that SANs in machine translation could learn word order mainly due to the PE, indicating that modeling cross-lingual information at position representation level may be informative. Thus, we propose a novel cross-lingual PE method to improve SANs.

\section{Conclusions and Future Work}
In this paper, we presented a novel cross-lingual position encoding to augment SANs by considering cross-lingual information (\textit{i.e.,} reordering indices) for the input sentence. We designed two strategies to integrate it into SANs. Experiments indicated that the proposed strategies consistently improve the translation performance. In the future, we plan to extend the cross-lingual position encoding to non-autoregressive MT~\cite{gu2018nonautoregressive} and unsupervised NMT~\cite{lample2018phrase}.

\section*{Acknowledgments}
This work was supported by Australian Research Council Projects FL-170100117. We are grateful to the anonymous reviewers and the area chair for their insightful comments and suggestions.

\bibliography{acl2020}
\bibliographystyle{acl_natbib}

\end{document}